\documentclass{article}
\usepackage[final]{neurips_2025}

\newcommand{\eat}[1]{}

\usepackage{latexsym}
\usepackage{amsfonts}
\usepackage{amsmath}
\usepackage{xcolor}
\usepackage{colortbl}
\usepackage{epsfig}
\usepackage{xspace}
\usepackage{graphicx}
\usepackage{subfigure}
\usepackage{paralist}
\usepackage{enumerate}
\usepackage{enumitem}
\usepackage{comment}
\usepackage{booktabs}
\usepackage{balance}
\usepackage{stmaryrd}
\usepackage{pifont}
\usepackage{listings}
\usepackage{array}
\usepackage{float}
\usepackage[flushleft]{threeparttable}

\usepackage{mathrsfs}
\usepackage{makecell}
\usepackage{xparse}
\usepackage{wrapfig}

\usepackage{epsfig}
\usepackage{multirow}
\usepackage{url}

\usepackage{multirow}
\usepackage{natbib}
\usepackage{graphicx}
\usepackage{fancyvrb}

\usepackage{listings}
\usepackage{framed}
\usepackage{xcolor}
\usepackage{color}
\colorlet{shadecolor}{gray!20}

\definecolor{shadecolor}{RGB}{220,220,220}

\newcommand{\stitle}[1]{\vspace{1ex}\noindent{\bf #1}}

\definecolor{inputcolor}{RGB}{255,139,35}
\definecolor{outputcolor}{RGB}{120,212,252}
\definecolor{embedcolor}{RGB}{254,127,156}
\definecolor{maskcolor}{RGB}{122,128,255}
\definecolor{ecolor}{RGB}{58,149,54}

\definecolor{highcolor}{RGB}{255,153,153}
\definecolor{midcolor}{RGB}{255,204,204}
\definecolor{lowcolor}{RGB}{204,229,255}

\usepackage{tikz}
\usetikzlibrary{shapes,snakes}
\usetikzlibrary{calc}

\usepackage[export]{adjustbox}

\definecolor{green}{RGB}{0,128,0}

\definecolor{yellow}{RGB}{255,200,18}

\newcommand{\bi}{\begin{itemize}}
\newcommand{\ei}{\end{itemize}}

\newcommand{\be}{\begin{enumerate}}
\newcommand{\ee}{\end{enumerate}}
\newcommand{\beqn}{\begin{eqnarray*}}
\newcommand{\eeqn}{\end{eqnarray*}}

\newcommand{\eg}{{\em e.g.,}\xspace}

\newcommand{\sys}{{DataPuzzle}\xspace}

\makeatletter
    \newcommand\figcaption{\def\@captype{figure}\caption}
    \newcommand\tabcaption{\def\@captype{table}\caption}
\makeatother

\tikzstyle{mybox} = [draw=black, fill=black!5, thick,
   rectangle, rounded corners, inner sep=0pt, inner ysep=6pt]
\tikzstyle{fancytitle} =[fill=black, text=white]

\NewDocumentCommand{\nan}{ mO{} }{\textcolor{blue}{\textsuperscript{\textit{Nan}}\textsf{\textbf{\small[#1]}}}}

\NewDocumentCommand{\zzx}{ mO{} }{\textcolor{yellow}{\textsuperscript{\textit{zzx}}\textsf{\textbf{\small[#1]}}}}

\title{DataPuzzle: Breaking Free from the Hallucinated Promise of LLMs in Data Analysis}

\author{%
Zhengxuan Zhang,
~Zhuowen Liang
~Yin Wu,
~Teng Lin,
~Yuyu Luo,
~Nan Tang \\
The Hong Kong University of Science and Technology (Guangzhou) \\
}

\makeatletter
\renewcommand{\@noticestring}{}
\makeatother

\begin{document}
\maketitle

\begin{abstract}

Large language models (LLMs) are increasingly applied to multi-modal data analysis---not necessarily because they offer the most precise answers, but because they provide fluent, flexible interfaces for interpreting complex inputs. Yet this fluency often conceals a deeper structural failure: the prevailing ``Prompt-to-Answer'' paradigm treats LLMs as black-box analysts, collapsing evidence, reasoning, and conclusions into a single, opaque response. The result is brittle, unverifiable, and frequently misleading. We argue for a fundamental shift: from generation to structured extraction, from monolithic prompts to modular, agent-based workflows. LLMs should not serve as oracles, but as collaborators---specialized in tasks like extraction, translation, and linkage---embedded within transparent workflows that enable step-by-step reasoning and verification. We propose \textbf{DataPuzzle}, a conceptual multi-agent framework that decomposes complex questions, structures information into interpretable forms (\eg tables, graphs), and coordinates agent roles to support transparent and verifiable analysis. This framework serves as an aspirational blueprint for restoring visibility and control in LLM-driven analytics---transforming opaque answers into traceable processes, and brittle fluency into accountable insight. This is not a marginal refinement; it is a call to reimagine how we build trustworthy, auditable analytic systems in the era of large language models. Structure is not a constraint---it is the path to clarity.

\end{abstract}

\section{Introduction}
\label{sec:intro}

\begin{center}
\emph{Somewhere along the way, we stopped building tools---and started summoning oracles.}
\end{center}

Prompt in, answer out—with no schema, no reasoning trace, just a shimmer of language draped over unknown logic. Large language models (LLMs) have rapidly become the default interface for data analysis—not because they are precise, but because they are \emph{conveniently fluent}~\cite{press2022measuring}. We now ask them to analyze spreadsheets, summarize medical records, compare visual patterns, and explain complex concepts across modalities~\cite{achiam2023gpt,li2024multimodal,gao2023pal}. It may feel like intelligence, but it is not.

Beneath this seductive fluency lies a dangerous confusion: \textbf{we have mistaken coherence for correctness, generation for understanding, and language for logic}~\cite{bender2021dangers}. The prevailing ``Prompt-to-Answer” paradigm treats LLMs as black-box analysts—give them access to data, craft the right prompt, and receive an insight~\cite{liu2023lost}. Yet beneath the polished surface lies no guarantee of \emph{truth}, \emph{process}, or \emph{control}~\cite{peng2023check}. The entire workflow becomes a theatrical performance—with no backstage.

\begin{figure}[t!]
\centering
\includegraphics[width=1\linewidth]{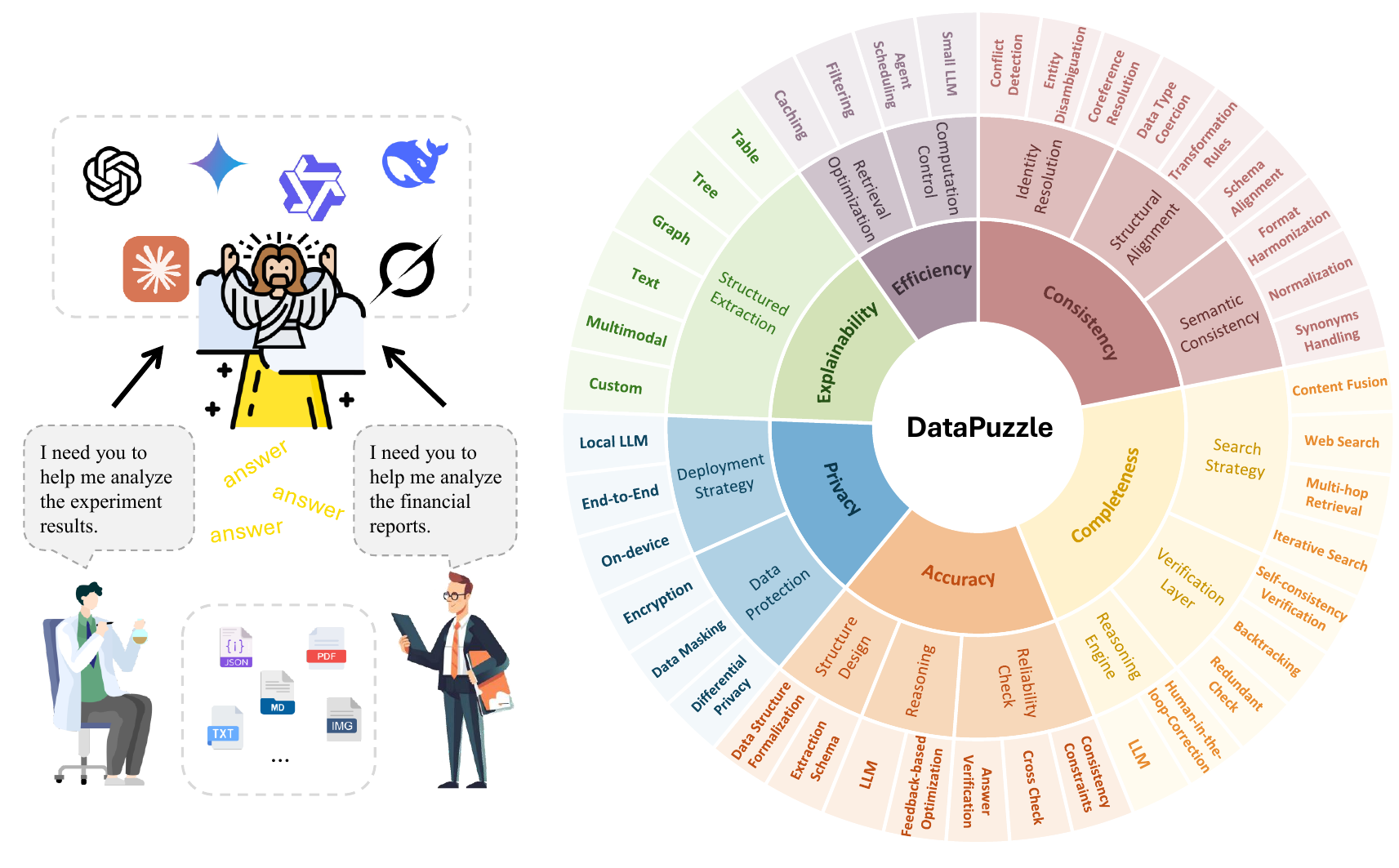} 
\caption{From Oracles to Systems: Six core dimensions (desiderata) for trustworthy multi-modal data analytics and potential research topics under each dimension.}
\label{fig:360}
\end{figure}

This illusion is beginning to fracture. Across applications, the same six fault lines appear again and again:

\be[leftmargin=1em]
\item \textbf{\textit{Not Explainable}}: 
	Multi-modal LLMs often face challenges in terms of explainability~\cite{dang2024explainable,qi2024sniffer}. When handling various data types such as text, images, and tables, these models do not provide a clear and understandable way of showing how they arrive at their results. Although they can process diverse data modalities, the internal processes and decision - making mechanisms are often opaque. 

\item \textbf{\textit{Low Accuracy}}: 
	Directly using LLMs to analyze a collection of data can lead to low accuracy~\cite{shrestha2025mathematical}. LLMs are not specifically engineered for accurate numerical or statistical analysis~\cite{zhularge,mirzadehgsm}. Their architecture and training are focused more on general language understanding and generation rather than precise quantitative computations. When confronted with complex tasks that require exact numerical calculations or in - depth domain - specific knowledge, their reasoning capabilities may fall short~\cite{verifai}. This can result in incorrect predictions, inaccurate classifications, or unreliable summaries, undermining the reliability of the analysis~\cite{DBLP:journals/debu/0001YZLF00H24}.

\item \textbf{\textit{Incomplete Data}}: 
	LLMs may not inherently identify or process all necessary information within or beyond a given dataset~\cite{ji2023survey}. They might miss key data points or fail to recognize missing information. In some cases, they may even attempt to ``hallucinate'' by generating plausible but incorrect information, which can introduce inaccuracies~\cite{huang2025survey}.

\item \textbf{\textit{Inconsistent Data}}: 
	When data from multiple sources is jointly used, inconsistencies may arise due to differences in formats, units, or even conflicting information~\cite{wang2024comprehensive}. LLMs may not inherently resolve these conflicts and could produce misleading results by synthesizing inconsistent data without flagging the discrepancies.

\item \textbf{\textit{Data Leakage}}: 
	One of the most significant concerns with using LLMs for data analytics is the potential exposure of private or sensitive data~\cite{xu2021privacy,pulido2020survey,DBLP:conf/icde/QinCTLLLZ22}. Sending private or confidential information (\eg personally identifiable information, financial data) to external LLMs poses risks of data breaches or misuse, especially if the model is hosted by third-party providers.

\item \textbf{\textit{Low Efficiency}}: 
	Using large-scale LLMs (\eg models with billions of parameters like DeepSeek-671B) for data analytics can result in high computational costs and latency~\cite{zhou2024survey,wang2024model}. These models require significant computational resources for inference, making them inefficient for real-time or large-scale data processing tasks.        
\ee
These fault lines are not just implementation bugs. They are \emph{symptoms} of a deeper structural failure: \textbf{we have no structure}. There is no separation between evidence and conclusion, no intermediate steps to verify, no representations to interrogate. These issues highlight several core desiderata—such as accuracy, completeness, consistency, efficiency, privacy, and overarching explainability—that trustworthy multi-modal data analytics systems must embody. Figure~\ref{fig:360} illustrates these critical dimensions and maps them to potential research challenges that need to be addressed.

\begin{center}
\textbf{We argue for a radical rethinking of LLM-based analytics. Not a tweak. A rupture.}
\end{center}

\stitle{From Generation to Extraction. From Oracles to Systems.}
LLMs are not analysts. They are not reasoners. They are not substitutes for logic or epistemology. But they \emph{are} powerful extractors, linkers, summarizers, and translators. The key is not to mask their limits, but to \textbf{reframe their role}.

We propose a new paradigm grounded in \textbf{structured extraction} and \textbf{step-by-step verification}. In this view, LLMs serve as collaborators—not decision-makers—within a larger analytic workflow. Analysis becomes a process of decomposition, grounding, verification, and composition. Answers are no longer hallucinated wholes, but mosaics of verifiable parts.

To realize this vision, we introduce \textbf{DataPuzzle}: a multi-agent framework for verifiable data analysis in the age of LLMs. DataPuzzle dismantles the Prompt-to-Answer illusion and rebuilds the workflow around three core principles:

\begin{itemize}[leftmargin=1em]
\item \textbf{Structure-first reasoning}, where intermediate representations (graphs, tables, event chains) scaffold analytic thought;
\item \textbf{Role-specialized agents}, each responsible for extracting, verifying, or reasoning over distinct aspects of the data;
\item \textbf{End-to-end transparency}, enabling inspection, auditing, and adaptation at every step.
\end{itemize}

\begin{figure*}[t!]
\centering
\includegraphics[width=1\linewidth]{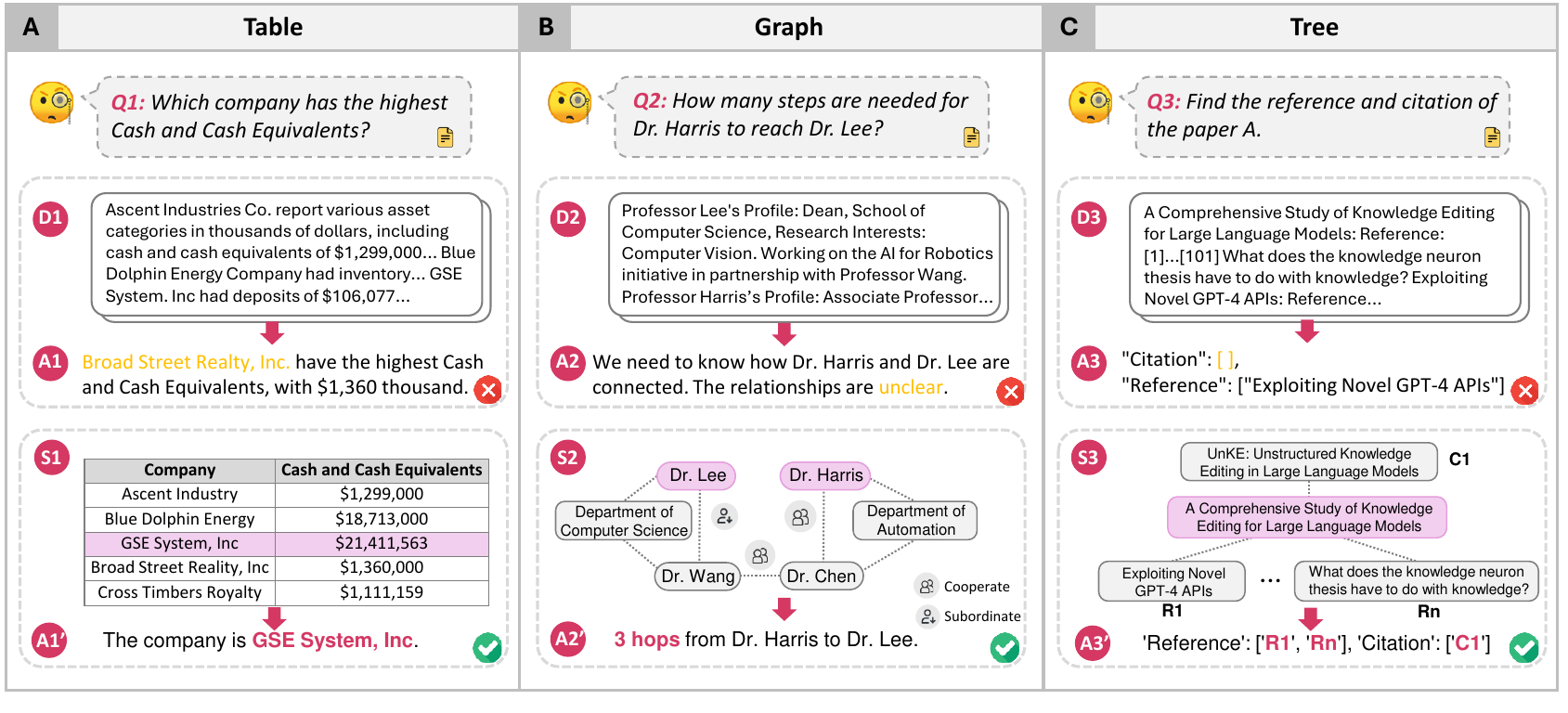}
\caption{Comparison of LLM Responses Before and After Data Structure Transformation. Here, $A_i$ is the answer obtained directly from the question $Q_i$ and the original document $D_i$, while $A_i'$ is the answer derived from the question $Q_i$ and the structured knowledge $S_i$ extracted from the original document $D_i$. Structure enhances verifiability and accuracy.}
\label{fig:mot}
\end{figure*}

What was once a one-shot prompt becomes a structured conversation between agents. What was once a guess becomes a traceable, interpretable computation. This is more than a technical intervention. It is a call to rethink what it means to \emph{analyze}, \emph{understand}, and \emph{trust} in the era of large language models. To illustrate the benefits of structure, Figure~\ref{fig:mot} contrasts LLM performance on raw documents ($D_i$) versus structured forms ($S_i$) derived from the same content. When data is organized into tables, graphs, or trees, LLMs can perform tasks such as comparison, pathfinding, and reference resolution more effectively. These structures expose the reasoning process, making it more transparent and verifiable, a cornerstone of \textbf{DataPuzzle}'s approach.

\stitle{Contributions.}
Our key contributions are:
\bi[leftmargin=2em]
\item We identify and discuss the critical fault lines in current LLM-based analytics and the corresponding desiderata for trustworthy and verifiable multi-modal data analysis, as highlighted by challenges in areas like accuracy, completeness, and explainability (Section~\ref{sec:problem}).
\item We propose \textbf{DataPuzzle}, an agentic framework employing an iterative think-extract-verify workflow, towards achieving explainable and verifiable multi-modal data analytics (Section~\ref{sec:framework}).
\item We further identify key open problems to guide future research in multi-modal data analytics (Section~\ref{sec:open_problem}).
\ei
\section{Problems and Desiderata}
\label{sec:problem}

\subsection{Problem of Multi-Modal Data Analytics}

Multi-modal data analytics aims to extract meaningful insights by integrating and analyzing data from diverse sources, such as text, images, audio, video, and sensor data. 

The \textbf{input} consists of diverse data sources in different modalities, such as text, images, audio, video, and sensor data, along with a natural language question that seeks to extract insights or answer a specific problem. These inputs are often heterogeneous in format, structure, and semantics, requiring integration and alignment for meaningful analysis.

The \textbf{output} is a unified response in the form of text (\eg summaries or explanations), tables (\eg structured data), charts (\eg visualizations), or a combination of these formats, designed to provide actionable insights that address the query while integrating information from multiple modalities.

\subsection{Desiderata}

\begin{figure}[t!]
    \centering
    \includegraphics[width=1\linewidth]{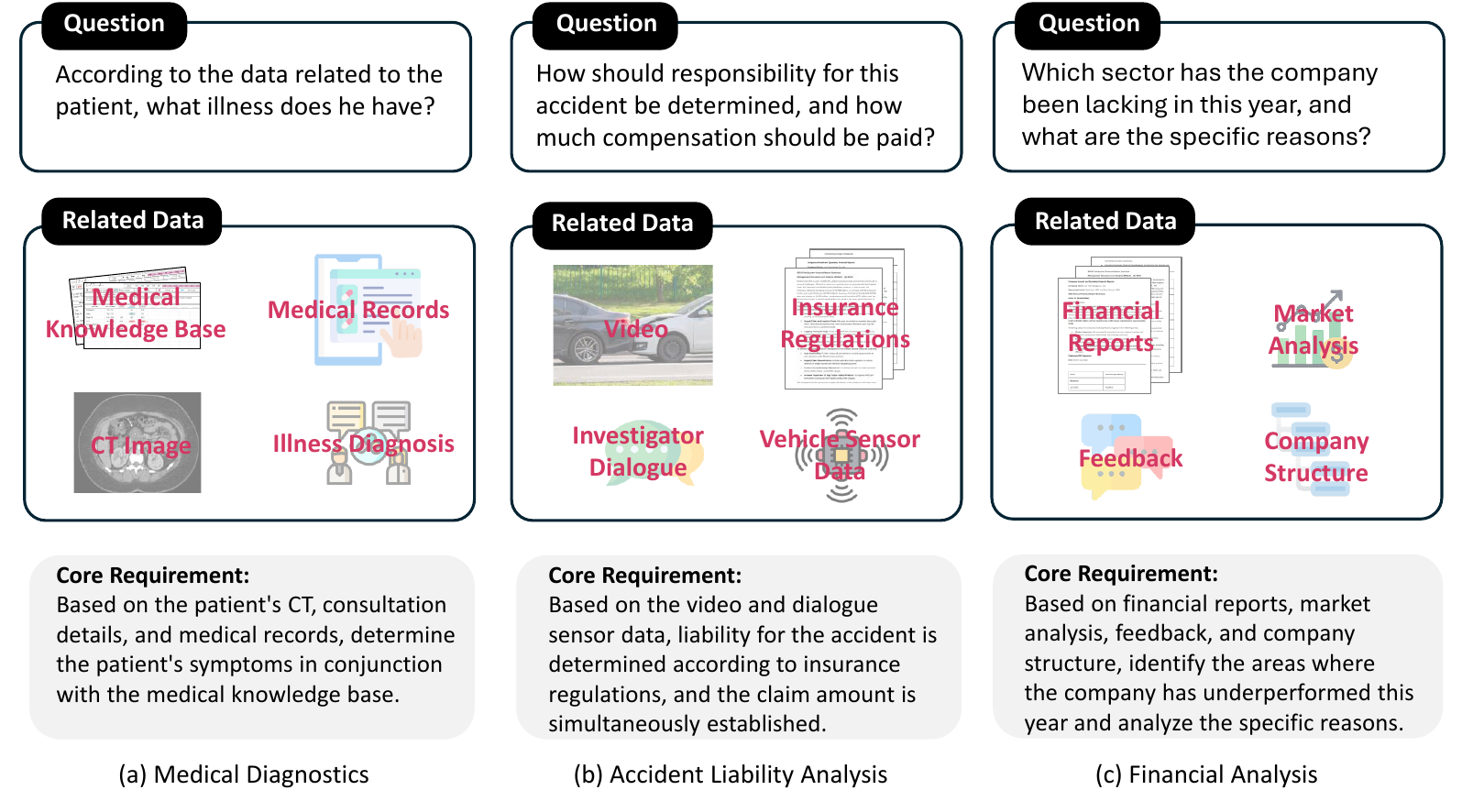}
    \caption{Illustrative three application scenarios, highlighting the multi-modal data and analytical requirements that motivate the system desiderata.}
    \label{fig:desiderata_case}
\end{figure}

To effectively support multi-modal data analytics across domains such as accident forensics, medical diagnostics, and business analysis, a system must satisfy several key desiderata. These requirements ensure that the system can integrate heterogeneous data, derive actionable insights, and remain robust in real-world high-stakes scenarios.

\stitle{Explainability.}
The system must provide clear and understandable explanations, particularly through structured data representation. This involves using explicit, meticulously curated data structures that are tailored to specific tasks. These structures allow for interpretability by clearly showing how data points and features influence the analysis, enhancing transparency and traceability.

\underline{\textit{Case Analysis:}} As shown in Figure~\ref{fig:desiderata_case}, in medical diagnostics this may involve mapping CT image features to symptom-diagnosis relations, helping clinicians understand how visual and textual evidence contributes to conclusions. In accident forensics, synchronized playback of sensor, video, and dialogue data allows investigators to trace causality and liability. In financial analysis, structured decomposition of indicators aligned with organizational units supports transparency in identifying weak sectors. Without explainability, users in all three domains may be unable to verify or trust the system’s outputs.

\stitle{High Accuracy.}
The system should achieve high precision, minimizing errors and ensuring reliability. It must be capable of accurate numerical and statistical computations, especially for tasks requiring precise quantitative insights, and incorporate domain-specific knowledge to enhance accuracy. 

\underline{\textit{Case Analysis:}} As illustrated in Figure~\ref{fig:desiderata_case}, in healthcare, errors in identifying clinical features from imaging or medical records could lead to misdiagnosis. In accident analysis, inaccuracies in synchronizing multi-source data may misattribute fault. In business settings, minor computational errors in financial metrics can mislead stakeholders. High accuracy is thus essential to ensure safe, fair, and impactful decisions across all domains.

\stitle{Completeness.}
The system must ensure that all relevant data points are identified and included in the analysis. This involves handling incomplete datasets through iterative search and verification to avoid missing critical information.

\underline{\textit{Case Analysis:}} As indicated in Figure~\ref{fig:desiderata_case}, medical diagnostics demand integration of CT scans, patient histories, and knowledge bases. Accident analysis relies on the fusion of video, sensor, and regulatory data. Business evaluations require financial reports, market signals, and internal feedback. In each case, omitting key data may obscure critical factors and undermine the validity of the findings.

\stitle{Consistency.}
The system should resolve inconsistencies across different data sources by aligning formats, units, and resolving conflicting information. This ensures that synthesized data is coherent and reliable.

\underline{\textit{Case Analysis:}} As shown in Figure~\ref{fig:desiderata_case}, in medicine, conflicting data across multiple tests or visits must be reconciled for a coherent diagnosis. Accident analysis needs alignment of measurement units and temporal information for credible reconstructions. Financial systems must normalize different reporting standards to support valid comparisons. Consistency ensures the integrity of integrated multi-modal outputs.

\stitle{Data Privacy.}
The system must protect sensitive data by minimizing exposure to external models. It should use locally deployable models and secure data handling practices to prevent data breaches or misuse.

\underline{\textit{Case Analysis:}} As shown in Figure~\ref{fig:desiderata_case}, in healthcare, compliance with standards like HIPAA or GDPR is mandatory to protect patient data. Accident data such as personal recordings or dialogues must be secured to prevent privacy breaches. In finance, internal documents and strategies are proprietary assets. Failure to enforce domain-specific privacy protections can lead to unauthorized data exposure, where sensitive user information—such as financial strategies or personal health records—may be inadvertently accessed or exploited by malicious actors.

\stitle{High Efficiency.}
The system should operate efficiently, minimizing computational costs and latency. This includes using fine-tuned small models and optimizing processes to handle large-scale data processing tasks effectively.

\underline{\textit{Case Analysis:}} As illustrated in Figure~\ref{fig:desiderata_case}, medical systems must produce low-latency results to support urgent decision-making. Accident claims require swift liability judgments to expedite recovery or arbitration. Financial dashboards must be efficiently generated from large-scale data to support agile strategy. Efficient system design—using lightweight models and scalable pipelines—is vital to meet these operational constraints.

\section{\sys: A Conceptual Blueprint for Trustworthy Analytics}
\label{sec:framework}

\begin{figure}
    \centering
    \includegraphics[width=1\linewidth]{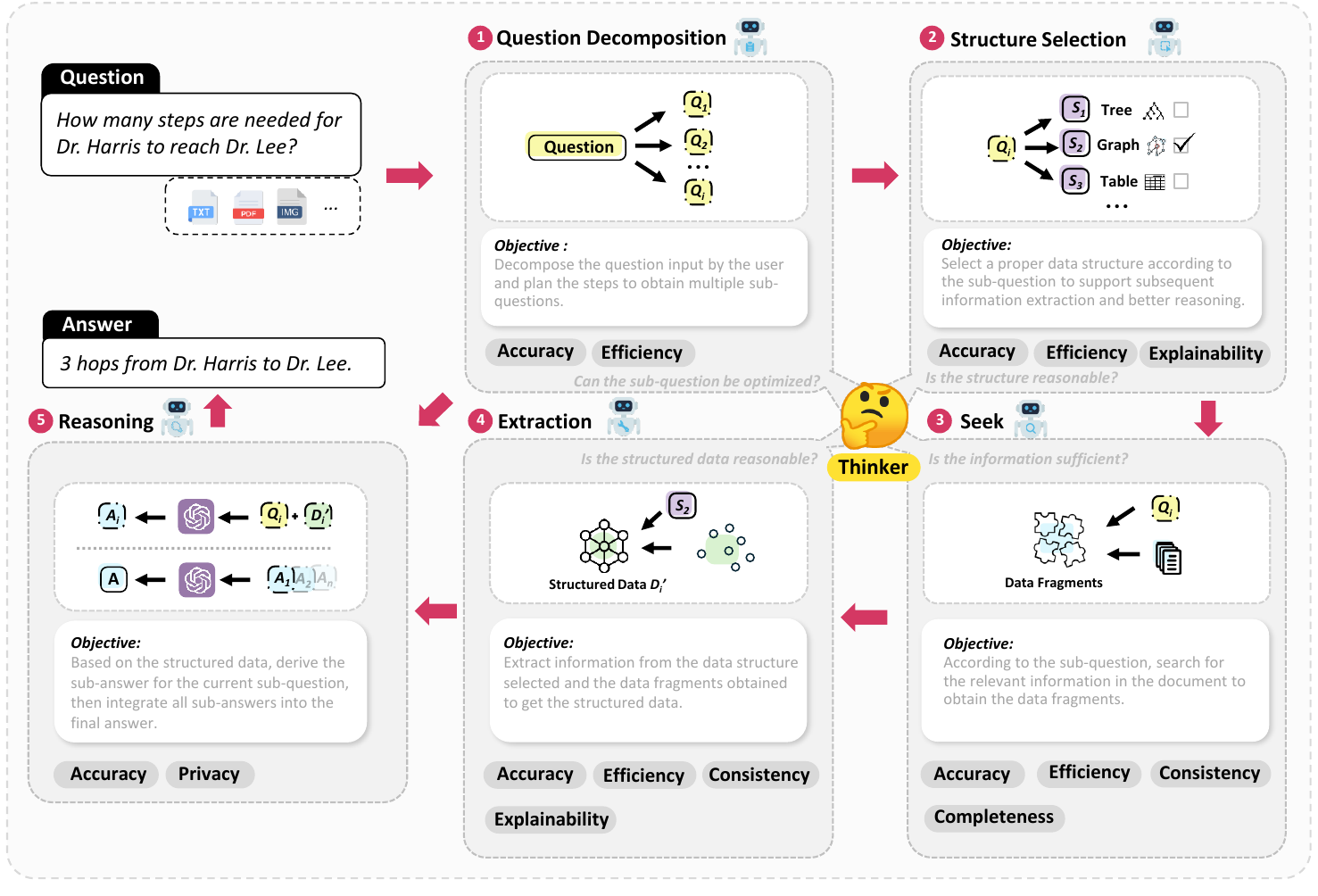}
    \vspace{-1em}
    \caption{Conceptual Blueprint of \sys for Trustworthy Analytics}
    \vspace{-1em}
    \label{fig:sys}
\end{figure}

To make our position more concrete, we outline a conceptual framework—\textbf{\sys}—that reimagines analytics pipelines around transparency, modular reasoning, and continuous oversight. It advocates a fundamental shift from black-box ``Prompt-to-Answer'' interactions toward structured, transparent, and verifiable workflows, as illustrated in Figure~\ref{fig:sys}. Rather than relying on monolithic models with limited introspection capabilities, \sys embraces modularity, structure-first reasoning, and iterative control through a central orchestrator, aiming to make the entire analytics process more explainable and trustworthy. In what follows, we describe the foundational philosophy, dynamic workflow, and the alignment of system design with the desiderata of trustworthy analytics.

\subsection{Foundational Philosophy: Structure, Modularity, and Oversight}

\stitle{Structure-First Reasoning.}
At the core of \sys is the principle of \textit{structure-first reasoning}. Raw multi-modal inputs—text, images, tables, or temporal data—are first abstracted into explicit intermediate representations such as graphs, trees, or tables. This transformation takes place early in the pipeline and serves a dual purpose: it abstracts away irrelevant details and surfaces task-relevant structures ($S_i$) upon which logic-based reasoning can be performed. These representations ($D_i'$) act as scaffolds that make each reasoning step inspectable, enabling not only better interpretability but also verifiability and correction when needed.

\stitle{Modular Agents.}
To manage the inherent complexity of multi-modal analytics, \sys envisions a modular conceptual architecture composed of role-specialized agents. Each module is designed to address a specific function—such as retrieval, extraction, or reasoning—potentially guided by task-specific prompts. This architectural separation ensures task modularity, supports local verification within each module, and simplifies orchestration.

\stitle{Meta-Level Oversight.}
A central component of the framework is a meta-level orchestrator, or the \textit{Thinker}, envisioned as a supervisory agent that monitors progress, evaluates intermediate results, and triggers replanning when needed. This layer of meta-level oversight transforms the system from a static workflow executor into a dynamic, adaptive reasoning entity. Importantly, this oversight is not simply reactive but designed for continuous improvement: it audits sub-question clarity, structure fit, data sufficiency, and logical consistency across phases. Together, structure-first processing, modular agents, and the Thinker’s feedback mechanisms form the foundation of a verifiable and adaptable analytics system.

\subsection{Iterative Workflow Under Thinker-Orchestrated Reasoning}

The operation of \sys unfolds as an iterative reasoning pipeline, initiated by a user query—often complex and multi-modal—and decomposed into manageable analytical tasks. Each stage of the pipeline is not only functionally distinct but also conceptually aligned with the overarching goal of trustworthy and auditable analytics.

\stitle{Decomposition \& Structure Selection.}
The process begins with \textbf{question decomposition}, in which the original query is broken down into a set of atomic sub-questions $\{q_1, q_2, \dots, q_n\}$. This step simplifies downstream processing and enables parallel reasoning tracks. For each sub-question $q_i$, an appropriate structure $S_i$ is selected—a decision influenced by the semantics of the sub-task and the modality of the data involved. For example, a table may be selected for comparison-based questions, while a graph might be chosen for relational or causal queries.

\stitle{Seek \& Extract.}
Once candidate structures are identified, the system transitions into a \textbf{Seek} phase, where relevant data fragments are retrieved from the available multi-modal sources. This retrieval is context-sensitive and may involve LLM-guided queries, vision-language grounding, or metadata-based filtering. Retrieved fragments are then passed to the \textbf{Extraction} stage, where they are converted into structured representations $D_i'$ compatible with the selected $S_i$. This step not only involves basic parsing but often demands information synthesis, normalization, and alignment—a task for which LLMs are employed under task-specific prompts and constrained scopes.

\stitle{Reasoning over Structures.}
Over the structured data $D_i'$, the \textbf{Reasoning} stage performs logic-based analysis to derive partial answers $A_i$. These results are then composed to form the final answer, often requiring aggregation, comparison, or contextual justification. 

\stitle{Thinker Oversight.}
Importantly, each of these phases is conceptually overseen by the \textbf{Thinker}. This agent continuously evaluates the fidelity of each step: whether the sub-questions are optimally phrased, the structures are well-matched to the task, the retrieved information is sufficient and diverse, and whether the extracted data is internally consistent and semantically valid. When needed, the Thinker may trigger a local revision (e.g., re-retrieval) or global re-planning (e.g., changing structure or re-decomposing the question), thus introducing a control loop for dynamic refinement.

\stitle{From Pipeline to Active Reasoner.} This iterative orchestration transforms \sys into an active reasoning system rather than a static pipeline executor. It allows the system to handle ambiguity, adjust to unexpected retrieval gaps, and recover from inconsistencies—capabilities essential for trustworthy decision-making in real-world analytical scenarios.

\subsection{Design-Driven Alignment with Desiderata of Trustworthy Analytics}

\sys builds trustworthiness into its design from the ground up. It integrates key goals—accuracy, completeness, consistency, efficiency, extensibility, and privacy—into every stage of the analytic process:

\textbf{Accuracy} can be supported by \textit{Question Decomposition} and \textit{Structure Selection}, which together allow complex user queries to be reframed as a series of tractable sub-questions, each aligned with an appropriate data representation. This structure-aware decomposition could ground subsequent reasoning steps in semantically faithful and logically coherent representations.

\textbf{Completeness} may be addressed during the \textit{Seek} process, where the system can actively assess whether retrieved information fragments sufficiently cover each sub-question. When necessary, iterative retrieval or refinement could be initiated to close information gaps, ensuring that no critical evidence is overlooked.

\textbf{Consistency} could be promoted through \textit{Seek} and \textit{Extraction}, where structured representations offer a foundation for normalization and conflict detection. By aligning retrieved content to predefined data structures, the system may prevent contradictions and promote coherence across disparate sources.

\textbf{Efficiency} is naturally facilitated by modular decomposition and structure-aware optimization across the pipeline. \textit{Question Decomposition}, \textit{Structure Selection}, and \textit{Extraction} can enable targeted computation, reuse of intermediate outputs, and lightweight agent orchestration---collectively reducing redundancy without compromising analytical depth.

\textbf{Explainability} is enhanced through \textit{Structure Selection} and \textit{Extraction}, which allow raw information to be grounded into interpretable forms such as tables, graphs, and trees. By structuring evidence in alignment with the analytic goal, the system enables transparent reasoning chains, clearer provenance tracking, and more faithful answer generation.

\textbf{Privacy} considerations can be embedded into \textit{Reasoning}, where operations may be performed over abstracted or locally processed data. Structure-aware privacy controls could support selective masking, modular isolation, and dynamic adaptation to sensitivity levels, minimizing unnecessary exposure.

\subsection{Vision: Toward Inherently Verifiable Analytics Systems}

By weaving together structure-first design, modular agents, and meta-level orchestration, \sys presents a blueprint for building analytics systems that are inherently trustworthy. Verification is not retrofitted as an evaluation step, but is instead deeply embedded in the fabric of the architecture. Each component contributes to a ``glass-box'' workflow where reasoning is inspectable, decisions are traceable, and the system retains the flexibility to adjust under uncertainty. As LLMs become central to data analysis tasks, frameworks like \sys will be essential for moving from intuition-based interactions to structured, controllable, and explainable reasoning—paving the way for the next generation of analytics systems.

\section{Open Problems}
\label{sec:open_problem}

\stitle{Information Extraction from New Data Modalities.}
Extracting actionable insights from emerging modalities (\eg 3D point clouds, real-time sensor streams) remains unsolved due to modality-specific representation gaps. For instance, aligning spatiotemporal sensor data with textual logs requires novel embedding spaces that preserve causal-temporal dependencies. Current methods (\eg contrastive learning) struggle with latent cross-modal invariants, such as mapping LiDAR geometry to natural language without losing structural fidelity, necessitating hybrid neuro-symbolic extractors. This challenge is further compounded by the need to handle data heterogeneity and noise inherent to real-world sensing environments, which complicates the design of robust and generalizable extraction frameworks. Moreover, the dynamic and streaming nature of some modalities calls for real-time extraction mechanisms capable of adapting on the fly, which remains largely unexplored. Future research must also focus on building unified models that can flexibly incorporate multiple emerging modalities without prohibitive computational costs.

\stitle{Information Extraction from New Domains.}
Domain shifts (\eg from finance to biomedicine) expose brittleness in schema induction and entity linking. A core challenge is domain-agnostic schema learning: automating the discovery of domain-specific ontologies (\eg legal clauses, genomic variants) without labeled data. This demands zero-shot transfer of extraction rules while avoiding semantic drift (\eg conflating ``risk'' in finance vs. healthcare), which current LLM-based adapters fail to address rigorously. Moreover, domain-specific jargon and implicit knowledge further hinder the accurate mapping between extracted entities and their true semantic roles, calling for more adaptive and interpretable domain transfer methods.

\stitle{Information Extraction for New Data Structures.}
Novel structures like hypergraphs (\eg multi-way scientific relationships) or topological maps lack extraction grammars. Key issues include defining minimal schema constraints for understudied structures (\eg cellular complexes in spatial data) and ensuring composability during merging. For example, how to extract a hypergraph from a mix of text and equations while preserving $n$-ary relations, without resorting to heuristics or manual curation. Addressing these issues is crucial for enabling downstream reasoning tasks that depend on high-fidelity relational representations across heterogeneous modalities and symbolic forms.

\stitle{Checking Completeness: Unknown Unknowns.}
Current completeness checks (\eg null counting, coverage metrics) fail to detect contextual omissions in multi-modal data. For instance, verifying whether a medical report’s table omits critical image findings requires reasoning about cross-modal entailments. Open problems include formalizing completeness certificates (proofs that all task-relevant data is extracted) via probabilistic logic, and detecting ``unknown unknowns'' through adversarial schema perturbations or information-theoretic coverage bounds. Developing such rigorous guarantees is essential to building trustworthy extraction systems that can flag potential blind spots before deployment in high-stakes domains.

\stitle{Causal Understanding in Multi-Modal Analysis.}
Moving from correlation to causation remains a foundational challenge in multi-modal data analytics. Existing systems struggle to uncover and reason about causal mechanisms that span heterogeneous modalities. Key open problems include: identifying cross-modal causal relationships (\eg how written guidelines affect observed behaviors in video); distinguishing genuine causal effects from spurious associations in noisy, high-dimensional data; and performing counterfactual reasoning across modalities with divergent structures and semantics. Addressing these challenges requires new causal inference frameworks that can operate over hybrid symbolic-neural representations while preserving interpretability and verifiability throughout the causal discovery and reasoning process.

\stitle{Human-Guided Reasoning over Structures.}
Most current pipelines stop at extraction and assume that downstream reasoning (\eg comparison, planning, inference) can be directly performed over the resulting structures. However, LLMs still struggle with goal-directed reasoning even when structured information is available. This calls for human-in-the-loop mechanisms that not only inspect but also steer the reasoning trajectory. Key open problems include: How can extracted graphs or tables be dynamically transformed to better support reasoning subgoals (\eg grouping, filtering, pathfinding)? How can the reasoning process be modularized to enable human feedback at intermediate steps—such as revising assumptions or validating intermediate conclusions? And how can the system align its reasoning steps more faithfully with the original analytic intent, avoiding irrelevant or misleading detours? These questions point to the need for a more deliberate interface between structure, reasoning, and human judgment—a planning layer that guides the traversal and transformation of structured data with both task goals and human cues in mind.

\section{Conclusion}
\label{sec:conclusion}

This paper challenges the dominant “Prompt-to-Answer” paradigm, which treats LLMs as oracles rather than tools. Such usage conflates coherence with correctness and lacks structure, traceability, and trust, leading to six major issues: unexplainability, low accuracy, poor data handling, inconsistency, data leakage, and inefficiency. 

We introduce DataPuzzle, a conceptual framework that repositions LLMs as collaborators within structured, agent-based workflows. By replacing one-shot generation with systematic extraction and verification, we transform answers from opaque outputs into verifiable composites. Our method rests on three principles: structure-first reasoning, role-specialized agents, and full transparency. This shift is more than technical—it redefines how we use LLMs for data analysis: not as all-knowing oracles, but as components of explainable, accurate, and trustworthy systems.

\stitle{Limitations.}
This paper presents a paradigm, not a complete solution. DataPuzzle is a conceptual and early-stage framework that prioritizes structural clarity over task-specific benchmarks. The agent-based design introduces computational overhead and coordination challenges that require careful system design—issues we leave for future work. Additionally, the effectiveness of our approach may vary across different domains and data types, requiring domain-specific adaptations.

\stitle{Ethics Statement.}
Our goal is to reduce opacity and risk in LLM-based analytics by promoting structure and verification. However, structured workflows are not foolproof. They must be designed with careful attention to avoid reinforcing bias, leaking sensitive data, or providing false confidence in results. The multi-agent approach requires thoughtful implementation to ensure that verification steps are robust and that the system remains transparent to users. Furthermore, as we develop more sophisticated analytical tools, we must remain vigilant about potential misuse and ensure that our systems promote equitable and ethical outcomes.

\bibliographystyle{abbrv}
\bibliography{refs/custom}

\end{document}